\documentclass{article}
\usepackage{times}
\usepackage{graphicx}
\usepackage{subfigure} 
\usepackage{natbib}
\usepackage{algorithm}
\usepackage{algpseudocode}
\usepackage{amsmath}
\usepackage{hyperref}
\usepackage{multirow}

\usepackage[seed=03492]{randomorder}   
\usepackage{arxiv} 

\newcommand{\vw}{\mathbf{w}}
\newcommand{\vv}{\mathbf{v}}
\newcommand{\vg}{\mathbf{g}}
\newcommand{\vzero}{\mathbf{0}}

\newcommand{\tra}{^{\mathsf{T}}}

\newcommand{\pderiv}[2]{\frac{\partial #1}{\partial #2}}

\newcommand{\primal}{elementary}
\newcommand{\Primal}{Elementary}

\newcommand{\hypers}{{\boldsymbol{\theta}}}
\newcommand{\params}{\vw}

\newcommand{\decay}{\gamma}
\newcommand{\decays}{{\boldsymbol{\decay}}}
\newcommand{\stepsize}{\alpha}
\newcommand{\stepsizes}{{\boldsymbol{\stepsize}}}
\newcommand{\gradparams}{\nabla_\params L(\params_t, \hypers, t)}

\newcommand\ourtitle{Gradient-based Hyperparameter Optimization through Reversible Learning}
\icmltitlerunning{\ourtitle} 

\newcommand\blfootnote[1]{%
  \begingroup
  \renewcommand\thefootnote{}\footnote{#1}%
  \addtocounter{footnote}{-1}%
  \endgroup
}

\begin{document} 

\twocolumn[
\icmltitle{\ourtitle}
\randomorder{
\icmlauthor{Dougal Maclaurin$^\dagger$}{maclaurin@physics.harvard.edu}
}{}
{\icmlauthor{David Duvenaud$^\dagger$}{dduvenaud@seas.harvard.edu}}
\icmlauthor{Ryan P. Adams}{rpa@seas.harvard.edu}

\icmlkeywords{hyperparameters, neural networks, reversible computation, automatic differentiation, machine learning, ICML}

\vskip 0.3in
]

\begin{abstract}
Tuning hyperparameters of learning algorithms is hard because gradients are usually unavailable.
We compute exact gradients of cross-validation performance with respect to all hyperparameters by chaining derivatives backwards through the \emph{entire training procedure}.
These gradients allow us to optimize thousands of hyperparameters, including step-size and momentum schedules, weight initialization distributions, richly parameterized regularization schemes, and neural network architectures.
We compute hyperparameter gradients by exactly reversing the dynamics of stochastic gradient descent with momentum.
\end{abstract} 

\section{Introduction}
\label{intro}

\blfootnote{$^\dagger$\explanationtext}
Machine learning systems abound with hyperparameters. These can be parameters
that control model complexity, such as $L_1$ and $L_2$ penalties, or parameters that
specify the learning procedure itself -- step sizes, momentum decay parameters
and initialization conditions. Choosing the best hyperparameters is both 
crucial and frustratingly difficult.

The current gold standard for hyperparameter selection is gradient-free model-based optimization~\cite{snoek2012practical, bergstra2011algorithms,
  BerYamCox13, HutHooLey11}.
Hyperparameters are chosen to optimize the validation loss after complete training of the model parameters.
These approaches have demonstrated that automatic tuning of hyperparameters can yield state-of-the-art performance.
However, in general they are not able to effectively optimize more than 10
to 20 hyperparameters.

Why not use gradients?
Reverse-mode differentiation allows gradients to be computed with a similar time
cost to the original objective function.
This approach is taken almost universally for optimization of \primal{}%
\footnote{Since this paper is about hyperparameters, we
  use ``\primal{}'' to unambiguously denote the other sort of parameter, the
  ``parameter-that-is-just-a-parameter-and-not-a-hyperparameter''.
}
parameters.
The problem with taking gradients with respect to hyperparameters is that computing the validation loss requires an inner loop of \primal{} optimization, which makes na\"ive reverse-mode differentiation infeasible from a memory perspective.
Section \ref{sec:hypergradients} describes this problem and proposes a solution, which is the main technical contribution of this paper.

Gaining access to gradients with respect to hyperparamters opens up a garden of
delights. Instead of straining to eliminate hyperparameters from our models, we
can embrace them, and richly hyperparameterize our models.
Just as having a high-dimensional \primal{} parameterization gives a
flexible model, having a high-dimensional hyperparameterization gives
flexibility over model classes, regularization, and training methods.
Section \ref{sec:experiments} explores these new opportunities.

\begin{figure}[t]
\begin{center}
\begin{tabular}{cc}
\renewcommand{\tabcolsep}{0pt}
\rotatebox{90}{\qquad \qquad \quad \small Training loss} &
\hspace{-.1in}\includegraphics[width=0.878\columnwidth, clip, trim=2cm 0cm 0cm 0cm]{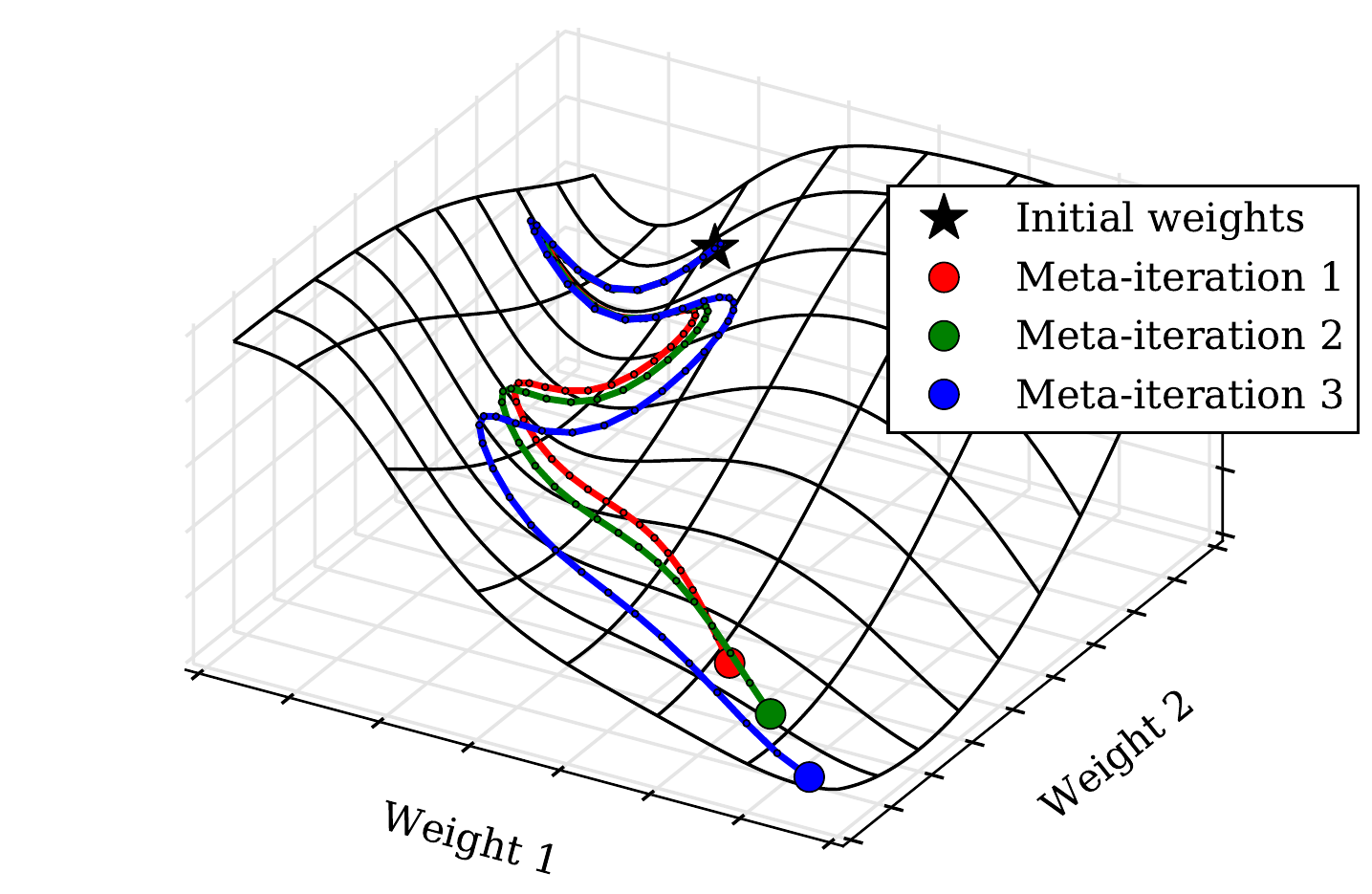}
\end{tabular}
\caption{Hyperparameter optimization by gradient descent.
Each meta-iteration runs an entire training run of stochastic gradient descent to optimize \primal{} parameters (weights 1 and 2).
Gradients of the validation loss with respect to hyperparameters are then computed by propagating gradients back through the \primal{} training iterations.
Hyperparameters (in this case, learning rate and momentum schedules) are then updated in the direction of this hypergradient.}
\label{fig:chaos}
\end{center}
\vskip -0.3in
\end{figure}

\subsection{Contributions}

\begin{itemize}
\item We give an algorithm that exactly reverses stochastic gradient descent with momentum to compute gradients with respect to all continuous training parameters.
\item We show how to efficiently store only the information needed to exactly reverse learning dynamics.
For example, when the momentum term is 0.9, this method reduces the memory requirements of reverse-mode differentiation of hyperparameters by a factor of 200.
\item We show that these gradients allow optimization of validation loss with respect to thousands of hyperparameters.
For example, we optimize fine-grained learning-rate schedules, per-layer initialization distributions of neural network parameters, per-input regularization schemes, and per-pixel data preprocessing.
\item We provide insight into learning procedures by examining optimized learning-rate schedules and initialization procedures, comparing them to standard advice in the literature.
\end{itemize}

\section{Hypergradients}
\label{sec:hypergradients}

Reverse-mode differentiation (RMD) has been an asset to the field of machine
learning~\citep{lecun1989backpropagation} (see the \ref{sec:appendix} for a refresher). The RMD method, known as
``backpropagation'' in the deep learning community, allows the gradient of a
scalar loss with respect to its parameters to be computed in a single backward
pass.
This increases the computational burden by only a factor of two over
evaluating the loss itself, regardless of the number of parameters.
Obtaining the same sort of information by either forward-mode
differentiation or brute force finite differences would require a separate pass
for each parameter and would make deep learning entirely infeasible.

Applying RMD to hyperparameter optimization was proposed by \citet{bengio2000gradient} and \citet{Autodiff14}, and applied to small problems by \citet{domke2012generic}.
However, the na\"ive approach fails for real-sized problems because
of memory constraints. RMD requires that intermediate variables be maintained
in memory for the reverse pass. Evaluating the validation loss
requires training the model, which may require many \primal{} iterations. Conventional
RMD stores this entire training trajectory, $\params_1
... \params_T$ in memory.  In large neural networks, the amount of
memory required to store the millions of parameters being trained is typically
close to the amount of physical RAM available~\cite{sequence2014}. If storing
the parameter vector takes $\sim$1GB, and the parameter vector is updated tens
of thousands of times (the number of mini batches times the number of
epochs) then storing the learning history is unmanageable even with
physical storage.

Imagine that we could exactly trace a training
procedure backwards, starting from the trained parameter values and working
back to the initial parameters. Then we could recompute the learning trajectory
on the fly during the reverse pass of RMD rather than storing it in memory. This is
not possible in general, but we will show that for the popular training
procedure of stochastic gradient descent with momentum, we can do
exactly this, storing a small number of auxiliary bits to handle
finite precision arithmetic.

\subsection{Reversible learning with exact arithmetic}
\label{sec:reversible learning}

Stochastic gradient descent (SGD) with momentum (Algorithm \ref{alg:sgd}) can be
seen as a physical simulation of a system moving through a series of fixed force
fields indexed by time $t$. With exact arithmetic this procedure is reversible. This lets us write Algorithm \ref{alg:reverse-sgd}, which reverses the steps in Algorithm
\ref{alg:sgd}, interleaved with computations of gradients. It outputs the
gradient of a function of the trained weights $f(\params)$ (such as the validation
loss) with respect to the initial weights $\params_1$, the learning-rate and momentum
schedules, and any other hyperparameters which affect training gradients.

\begin{algorithm}
   \caption{Stochastic gradient descent with momentum}
   \label{alg:sgd}
\begin{algorithmic}[1]
   \State {\bfseries input:} initial $\params_1$, decays $\decays$, learning rates $\stepsizes$, loss function $L(\params, \hypers, t)$
   \State initialize $\vv_1 = \vzero$
   \For{$t=1$ {\bfseries to} $T$}
   \State $\vg_t = \gradparams$ \Comment{evaluate gradient}
   \State $\vv_{t+1} = \decay_t \vv_t - (1 - \decay_t) \vg_t$ \Comment{update velocity}  \label{step:update velocity}
   \State $\params_{t+1} = \params_t + \stepsize_t \vv_t$ \Comment{update position} \label{step:update position}
   \EndFor
   \State \textbf{output} trained parameters $\params_T$
\end{algorithmic}
\end{algorithm}
\begin{algorithm}
   \caption{Reverse-mode differentiation of SGD}
   \label{alg:reverse-sgd}
\begin{algorithmic}[1]
   \State {\bfseries input:} $\params_T$, $\vv_T$, $\decays$, $\stepsizes$, train loss $L(\params, \hypers, t)$, loss $f(\params)$
   \State initialize $d\vv = \vzero$, $d\hypers = \vzero$, $d\stepsize_t = \vzero$, $d\decay = \vzero$
   \State initialize $d\params = \nabla_\params f(\params_T)$
   \For{$t=T$ {\bfseries counting down to} $1$}
   \State $d\stepsize_t = d\params\tra \vv_t$
   \State $\params_{t-1} = \params_t - \stepsize_t \vv_t$ \label{step:reverse-position}
   \vspace{-0.95\baselineskip}
   \State $\vg_t = \gradparams$ \label{step:reverse-gradient}
   \hfill \scalebox{1.1}{\Bigg\}} \vspace{-\baselineskip} \begin{minipage}{2.5cm} exactly reverse \\ gradient descent \\ operations \strut \end{minipage}
   \State $\vv_{t-1} = [\vv_t + (1 - \decay_t) \vg_t] / \decay_t$ \label{step:reverse-velocity}
   \State $d\vv = d\vv + \stepsize_t d\params$
   \State $d\decay_t = d\vv\tra (\vv_t + \vg_t)$
   \State $d\params = d\params - (1 - \decay_t) d\vv \nabla_\params \gradparams$ \label{line:hvp1}
   \State $d\hypers = d\hypers - (1 - \decay_t) d\vv \nabla_\hypers \gradparams$ \label{line:hvp2}
   \State $d\vv = \decay_t d\vv$
   \EndFor
   \State \textbf{output} gradient of $f(\params_T)$ w.r.t $\params_1$, $\vv_1$, $\decays$, $\stepsizes$ and $\hypers$
\end{algorithmic}
\end{algorithm}
Computations of steps \ref{line:hvp1} and \ref{line:hvp2} both require a
Hessian-vector product, but these can be computed exactly by applying RMD to the
dot product of the gradient with a vector \citep{pearlmutter1994fast}.  Thus the
time complexity of reverse SGD is $\mathcal{O}(T)$, the same as forward SGD.

\subsection{Reversible learning with finite precision arithmetic}

In practice, Algorithm \ref{alg:reverse-sgd} fails utterly due to finite
numerical precision. The problem is the momentum decay term~$\decay$.
Every time we apply step \ref{step:reverse-velocity} to reduce the velocity, we
lose information. Assuming we are using a fixed-point representation,
\footnote{We assume fixed-point representation to simplify the discussion (and
  the implementation). 
  \citet{courbariaux2014low} show that fixed-point arithmetic is sufficient to train deep networks.
  Floating point representation doesn't fix the problem, it
  just defers the loss of information from the division step to the addition step.}
each multiplication by $\decay < 1$ shifts bits to the right, destroying the
least significant bits. This is more than a pedantic concern. Attempting to
carry out the reverse training requires repeated multiplication by $1/\decay$. 
Errors accumulate exponentially, and the reversed learning procedure ends far
from the initial point (and usually overflows). Do we need $\decay < 1$?
Unfortunately we do. $\decay > 1$ results in unstable dynamics, and
$\decay = 1$, recovers the leapfrog integrator~\citep{leapfrog1995}, a perfectly reversible set of dynamics, but one that does not converge.

This problem is quite a deep one: optimization necessarily discards information.
Ideally, optimization maps all initializations to the same optimum,
a many-to-one mapping with no hope of inversion.
Put another way, optimization moves a system from a high-entropy initial state
to a low-entropy (hopefully zero entropy) optimized final state.

It is interesting to consider the analogy with physical dynamics. The $\decay$
term is analogous to a drag term in the simulation of Hamiltonian dynamics.
Having $\decay < 1$ corresponds to \emph{dissipative} dynamics which generates
heat, increases the entropy of the environment and is not therefore not
reversible. But we must have dissipation in order for our system to converge to equilibrium.

If we want to reverse the dynamics, there is no choice but to store the extra
bits discarded by the $\decay$ operation. But we can at least try to
be parsimonious about the number of extra bits we store. This is what the next
section addresses.

\subsection{Optimal storage of discarded entropy}
\label{sec:reversible computation}

This section gives the technical details of how to efficiently store the information discarded each time the momentum decay
operation (Step \ref{step:reverse-velocity}) is applied.

If $\decay = 0.5$, we can simply store the
single bit that falls off at each iteration, and if $\decay = 0.25$ we could
store two bits. But for fine-grained control over $\decay$ we need a way to store the information lost when we multiply by, say, $\decay = 0.9$, which will be less than one bit on average. Here we give a procedure which achieves exactly
this.

We represent the velocity $\vv$ and parameter $\params$ vectors with 64-bit integers. With an implied radix point this can be a fixed-point
representation of the reals. We represent $\decay$ as a rational number,
$n/d$. When we divide each $v$ by $d$ we use integer division. In order to be able to
reverse the process we just need to store the remainder, $v$ modulo $s$, in some
``information buffer'', $B$. If $B$ were an integer and $n = 2$, the remainder $r$ would just be a single bit, and we could store it in $B$ by left-shifting $B$'s bits and adding $r$. For arbitrary $n$, we can do the base-$n$ analogue of this operation: multiply $B$ by $n$ and add $r$.
Eventually, $B$ will overflow. We need a way to either detect
this, store the bits, and start a fresh integer, or else we can just use an
arbitrary size integer that grows as needed. (Python's ``long'' integer type
supports this). This procedure allows division by $n$ while storing the
remainder in $\log_2(n)$ bits on average.

When we multiply by the numerator of $n/d$ we don't need to store anything
extra, since integer division will bring us back to exactly the same point
anyway. But the procedure as it stands would store three bits when $\decay = 7/8$,
whereas it should store less than one ($\log_2 (8/7) = 0.19$). Our solution is the following: when we multiply $v$ by $n$, there is an opportunity to add a
nonnegative integer smaller than $n$ to the result without affecting the reverse
process (integer division by $n$). We can get such an integer from the
information buffer by dividing it by $n$ and recording $B$ modulo $n$. We are using
the velocity $v$ as an information buffer itself! Algorithm
\ref{alg:reversible-mult} illustrates the entire process.

\begin{algorithm}
   \caption{Exactly reversible multiplication by a ratio}
   \label{alg:reversible-mult}
\begin{algorithmic}[1]
   \State {\bfseries Input:} Information buffer $i$, value $c$, ratio $n / d$
   \State $i = i \times d$ \Comment{make room for new digit}               \label{step:f1}
   \State $i = i + (c \! \mod d)$ \Comment{store digit lost by division}   \label{step:f2}
   \State $c = c \div d$ \Comment{divide by denominator}                   \label{step:f3}
   \State $c = c \times n$ \Comment{multiply by numerator}                 \label{step:b1}
   \State $c = c +  (i \! \mod n)$ \Comment{add digit from buffer}         \label{step:b2}
   \State $i = i \div n$ \Comment{shorten information buffer}              \label{step:b3}
   \State \textbf{return} updated buffer $i$, updated value $c$
\end{algorithmic}
\end{algorithm}

We could also have used an arithmetic coding scheme for our information buffer~\citep[Chapter 6]{mackay2003information}.
How much does this procedure save us? When $\decay =  0.98$, we will have
to store only $0.029$ bits on average. Compared to storing a new 32-bit integer or
floating-point number at each iteration, this reduces memory requirements by a factor of one thousand.

The standard way to save memory in RMD is checkpointing. 
Checkpointing stores the entire parameter vector on only a fraction of the training steps, and recomputes the missing steps of the training procedure (forwards) as needed during the backward pass.
However, this would require too much memory to be practical for large neural nets trained
for thousands of minibatches.

\section{Experiments}
\label{sec:experiments}


In typical machine learning applications, only a few hyperparameters (less than 20) are optimized.
Since each experiment only yields a single number (the validation loss), the search rapidly becomes more difficult as the dimension of the hyperparameter vector increases.
In contrast, when hypergradients are available, the amount of information gained from each training run grows along with the number of hyperparameters, allowing us to optimize thousands of hyperparameters. 
How can we take advantage of this new ability?

This section shows several proof-of-concept experiments in which we can more richly parameterize training and regularization schemes in ways that would have been previously impractical to optimize.

\subsection{Gradient-based optimization of gradient-based optimization}
\label{sec:schedule experiments}
Modern neural net training procedures often employ various heuristics to set learning rate schedules, or set their shape using one or two hyperparameters set by cross-validation \cite{dahl2014multi, sutskever2013importance}.
These schedule choices are supported by a mixture of intuition, arguments about the shape of the objective function, and empirical tuning.

To more directly shed light on good learning rate schedules, we jointly optimized separate learning rates for \emph{every single learning iteration} of training of a deep neural network, as well as separately for weights and biases in each layer.
Each meta-iteration trained a network for 100 iterations of SGD, meaning that the learning rate schedules were specified by 800 hyperparameters (100 iterations $\times$ 4 layers $\times$ 2 types of parameters).
To avoid learning an optimization schedule that depended on the quirks of a particular random initialization, each evaluation of hypergradients used a different random seed.
These random seeds were used both to initialize network weights and to choose mini batches.
The network was trained on 10,000 examples of MNIST, and had 4 layers, of sizes 784, 50, 50, and 50.

Because learning schedules can implicitly regularize networks~\cite{erhan2010does}, for example by enforcing early stopping, for this experiment we optimized the learning rate schedules on the training error rather than on the validation set error.
\begin{figure}[h]
\vskip 0.1in
\begin{center}
Optimized learning rate schedule \\
\includegraphics[width=\columnwidth]{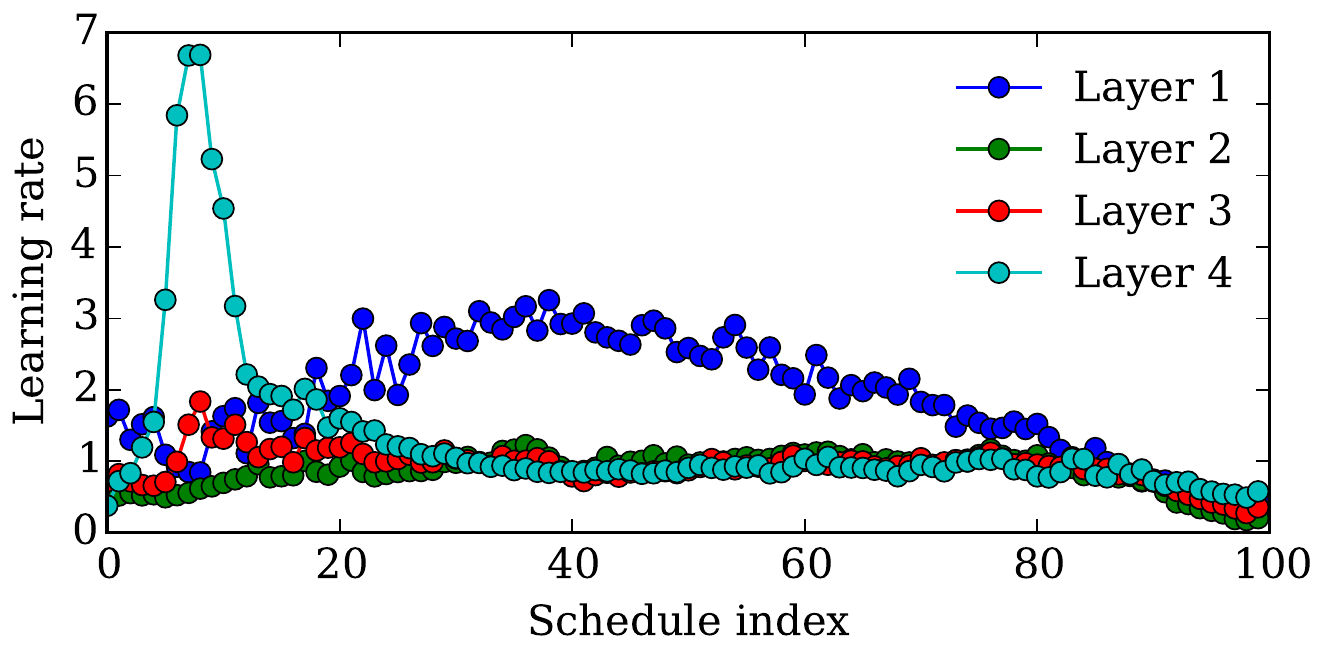}
\vskip -0.1in
\caption{A learning-rate training schedule for the weights in each layer of a neural network, optimized by hypergradient descent.
The optimized schedule starts by taking large steps only in the topmost layer, then takes larger steps in the first layer.
All layers take smaller step sizes in the last 10 iterations.
Not shown are the schedules for the biases or the momentum, which showed less structure.}
\label{fig:optimal schedule}
\end{center}
\vskip -0.1in
\end{figure} 
Figure \ref{fig:optimal schedule} shows the results of optimizing learning rate schedules separately for each layer of a deep neural network.
When Bayesian optimization was used to choose a fixed learning rate for all layers and iterations, it chose a learning rate of 2.4.

\begin{figure}[t]
\begin{center}
\begin{tabular}{cc}
 \Primal{} learning curves & Meta-learning curve \\
\hspace{-1em}\includegraphics[width=0.5\columnwidth, height=0.5\columnwidth]{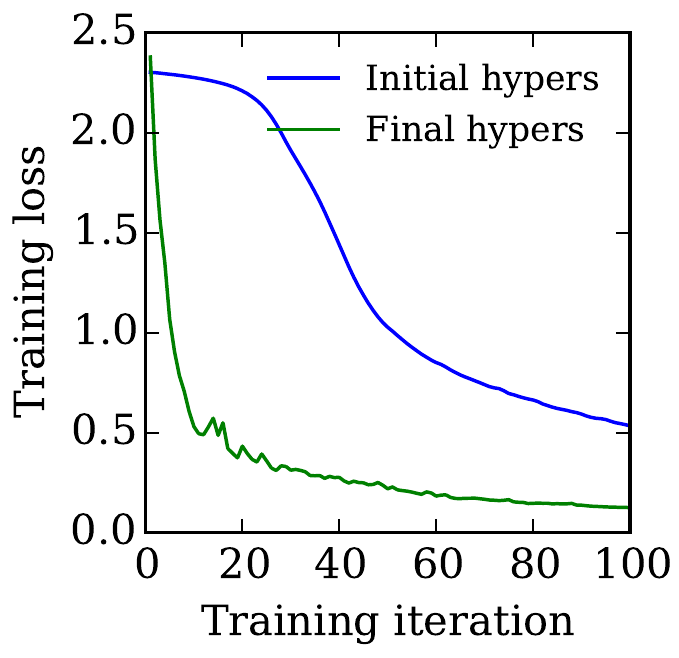} &
\hspace{-1em}\includegraphics[width=0.5\columnwidth, height=0.5\columnwidth]{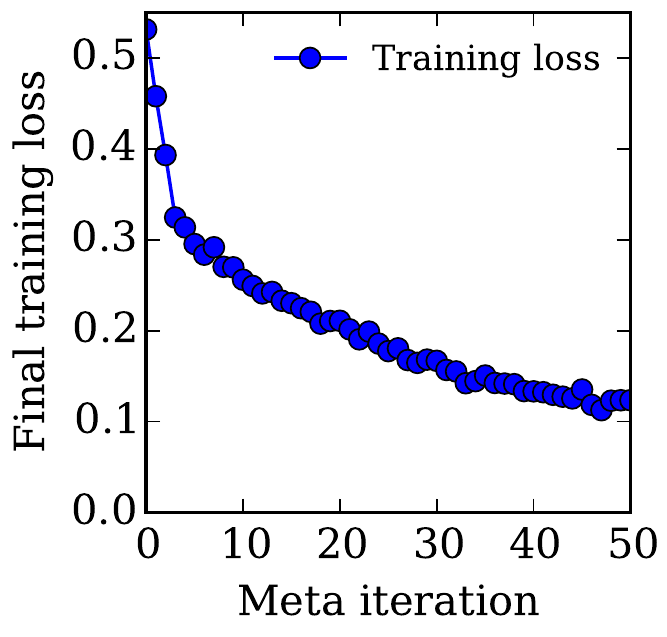}
\end{tabular}
\vskip -0.1in
\caption{\Primal{} and meta-learning curves.
The meta-learning curve shows the training loss at the end of each \primal{} iteration.}
\label{fig:learning curves}
\end{center}
\vskip -0.2in
\end{figure}

\paragraph{Meta-optimization strategies}
We experimented with several standard stochastic optimization methods for meta-optimization, including SGD, RMSprop~\citep{Tieleman2012}, and minibatch conjugate gradients.
The results in this section used Adam~\citep{Adam14}, a variant of RMSprop that includes momentum.
We typically ran for 50 meta-iterations, and used a meta-step size of 0.04.
Figure \ref{fig:learning curves} shows the \primal{} and meta-learning curves that generated the hyperparameters shown in Figure \ref{fig:optimal schedule}.

\begin{figure}[t]
\vskip 0.1in
\begin{center}
Hypergradient at first meta-iteration\\
\includegraphics[width=\columnwidth]{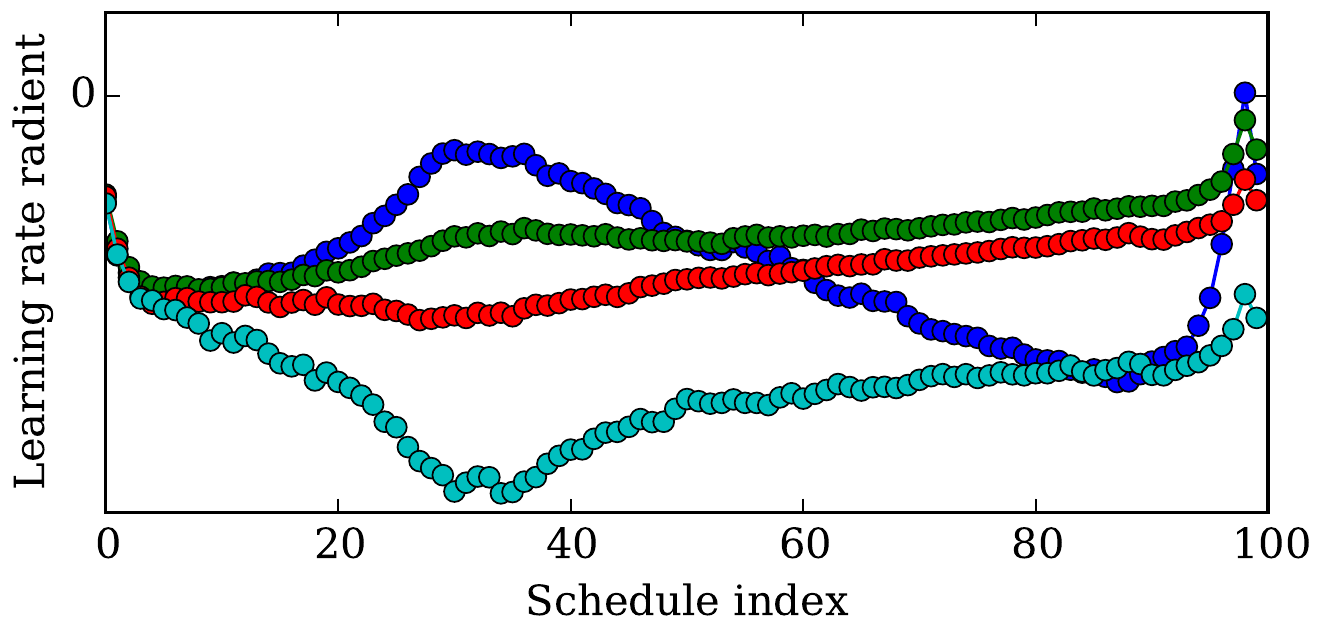}
\vskip -0.1in
\caption{The initial gradient of the cross-validation loss with respect to the training schedule, averaged over 100 random weight initializations and mini batches.
Colors correspond to the same layers as in Figure \ref{fig:optimal schedule}.}
\label{fig:smoothed gradient}
\end{center}
\vskip -0.2in
\end{figure}

\paragraph{How smooth are hypergradients?}
To demonstrate that the hypergradients are smooth with respect to time steps in the training schedule, Figure \ref{fig:smoothed gradient} shows the hypergradient with respect to the step size training schedule at the beginning of training, averaged over 100 random seeds.

\begin{figure}[t]
\vskip 0.2in
\begin{center}
\begin{tabular}{cc}
 Biases & Weights \\
\hspace{-1em}\includegraphics[width=0.5\columnwidth, height=0.5\columnwidth]{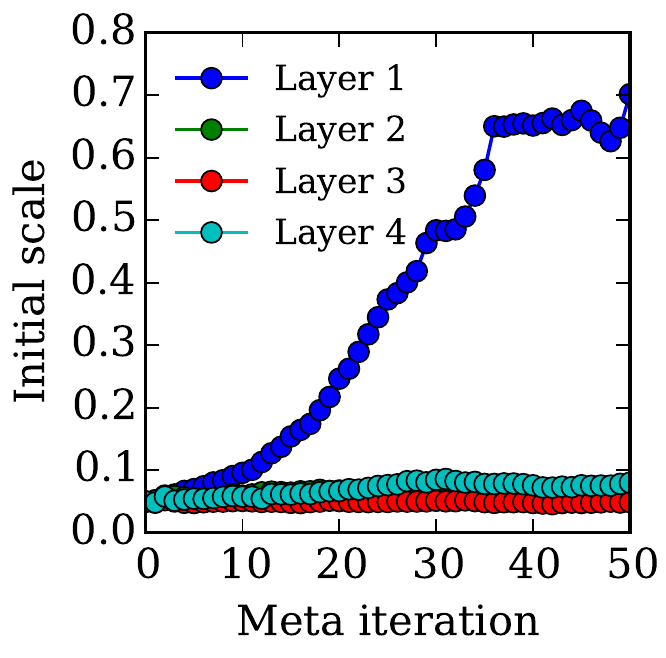} &
\hspace{-1em}\includegraphics[width=0.5\columnwidth, height=0.5\columnwidth]{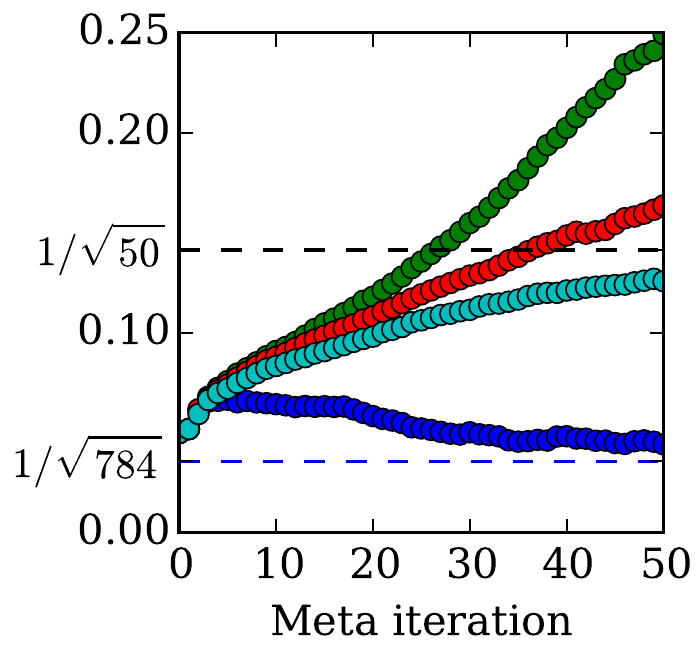}
\end{tabular}
\caption{Meta-learning curves for the initialization scales of each layer in a 4-layer deep neural network.
\emph{Left:} Initialization scales for biases.
\emph{Right:} Initialization scales for weights.
Dashed lines show a heuristic which gives an average total activation of 1.
For the first layer it is ($1/\sqrt{784}$) and for subsequent layers ($1/\sqrt{50}$).}
\label{fig:nn weight init scales}
\end{center}
\vskip -0.2in
\end{figure}

\paragraph{Optimizing weight initialization scales}
We optimized a separate weight initialization scale hyperparameter for each type of parameter (weights and biases) in each layer - a total of 8 hyperparameters.
Results are shown in Figure \ref{fig:nn weight init scales}.

Interestingly, the initialization scale chosen for the first layer weights matches a heuristic which says to choose an initialization scale of $1/\sqrt{N}$, where $N$ is the number of weights in the layer.

\subsection{Optimizing regularization parameters}
\label{sec:optimizing regularization}
Regularization is often important for generalization performance.
Typically, a single parameter controls a single $L_2$ norm or sparsity penalty on the entire parameter vector of a neural network.
Because different types of parameters in different layers play different roles, it is reasonable to suspect that separate regularization hyperparameter for each parameter type would improve performance.
Indeed, \citet{snoek2012practical} optimized separate regularization parameters for each layer in a neural network, and found that it improved performance.

We can take this idea even further, and introduce a separate regularization penalty for each individual parameter in a neural network.
We use a simple model as an example -- logistic regression, which can be seen as a neural network without a hidden layer.
We choose this model because every weight corresponds to an input-pixel and output-label pair, meaning that these 7,840 hyperparameters might be relatively interpretable.
\begin{figure}[t]
\begin{center}
\includegraphics[width=\columnwidth]{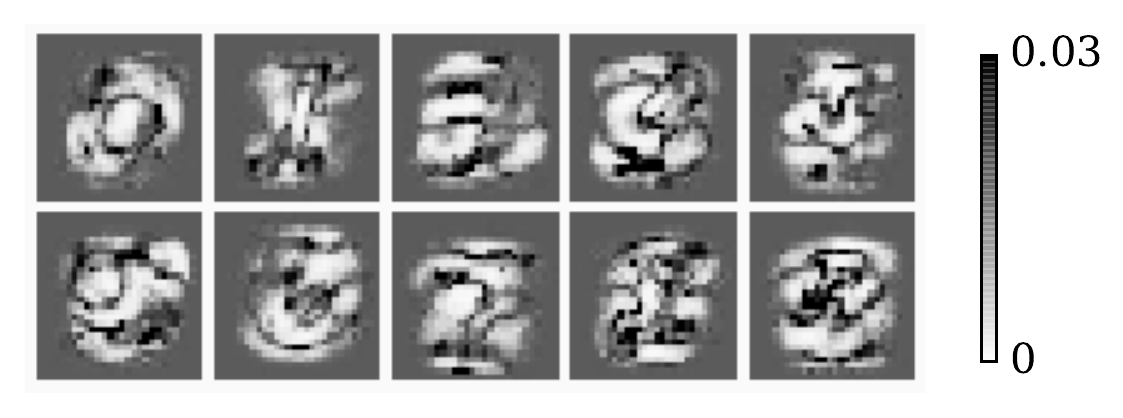}
\vspace{-2em}
\caption{Optimized $L_2$ regularization hyperparameters for each weight in a logistic regression  trained on MNIST.
The weights corresponding to each output label (0 through 9 respectively) have been rendered separately.
High values (black) indicate strong regularization.}%
\label{fig:logistic ard}%
\end{center}
\vskip -0.15in
\end{figure} 
Figure \ref{fig:logistic ard} shows a set of regularization hyperparameters learned for a logistic regression network.
Because each parameter corresponds to a particular input, this regularization scheme could be seen as a generalization of automatic relevance determination~\citep{mackay1994automatic}.

\subsection{Optimizing training data}

We can use Algorithm \ref{alg:reverse-sgd} to take the gradient with respect to any parameter the training procedure depends on.
This includes the training data, which can be viewed as just another set of hyperparameters.
By chaining gradients through transformations of the data, we can compute gradients of the validation objective with respect to data preprocessing, weighting, or augmentation procedures.

\begin{figure}[b]
\begin{center}
\includegraphics[width=\columnwidth]{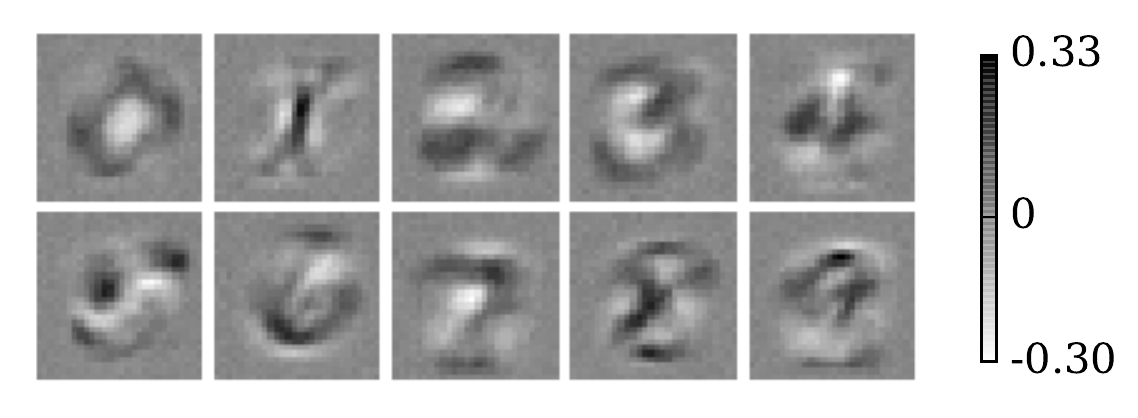}
\vskip -0.1in
\caption{A dataset generated purely through meta-learning.
Each pixel is treated as a hyperparameter, which are all optimized to maximize validation-set performance.
Training labels are fixed in order from 0 to 9.
Some optimal pixel values are negative.}
\label{fig:fake data}
\end{center}
\vskip -0.1in
\end{figure}
We demonstrate a simple proof-of-concept where an \emph{entire training set} is learned by gradient descent, starting from blank images.
Figure \ref{fig:fake data} shows a training set, the pixels of which were optimized to improve performance on a validation set of 10,000 examples from MNIST.
We optimized 10 training examples, each having a different fixed label, again from 0 to 9 respectively.
Learning the labels of a larger training set might shed light on which classes are difficult to distinguish and so require more examples.

\subsection{Optimizing initial parameters}
The last remaining parameter to SGD is the initial parameter vector.
Treating this vector as a hyperparameter blurs the distinction between learning and meta-learning.
In the extreme case where all \primal{} learning rates are set to zero, the training set ceases to matter and the meta-learning procedure exactly reduces to \primal{} learning on the validation set.
Due to philosophical vertigo, we chose not to optimize the initial parameter vector.

\subsection{Learning continuously parameterized architetures}

Many of the notable successes in deep learning have come from novel
architectures adapted to particular domains: convolutional neural nets,
recurrent neural nets and multitask neural nets. We can think of these
architectures as hard constraints that force particular weights to be zero and
tie particular pairs of weights together. By softening these hard architectural
constraints we can form continuous (but very high-dimensional) parameterizations
of architecture. Having access to hypergradients makes learning these
softened architectures feasible.

We illustrate this ``architecture learning'' with a multitask learning
problem, the Omniglot data set \citep{Omniglot}. This data set consists of 28x28
pixel greyscale images of characters from 50 alphabets with up to 55 characters in
each alphabet but only 15 examples of each character. Rather than learning a
separate neural net for each alphabet, a multitask approach would be for all the
neural nets to share a single first layer, pooling statistical strength to learn
generic Gabor-like filters, while maintaining separate higher layers specific to
each alphabet.

We can parameterize any architecture based on weight tying or weight absence
with a pairwise quadratic penalty on the weights, $\vw^T A \vw$, where $A$ is a
number-of-weights by number-of-weights matrix. Learning this enormous matrix is
clearly infeasible but we can implicitly build such a matrix from lower
dimensional structures of manageable size.

For the Omniglot problem, we learn a penalty for each alphabet pair, separately
for each neural net layer. Thus, for ten three-layer neural networks,
the penalty matrix $A$ is fully described by three ten-by-ten matrices. An
architecture with fully independent nets for each alphabet corresponds to three
diagonal matrices while an architecture with a mutual lower layer corresponds to
two diagonal matrices for the upper layers and a matrix of all ones
for the lowest layer (Figure \ref{fig:omniglot_results}).

\begin{figure}[h!]
\begin{center}
\begin{tabular}{cc}
\hspace{-3mm}\rotatebox{90}{\qquad Rotated \qquad \quad Original} & 
\hspace{-3mm}\includegraphics[width=0.85\columnwidth]{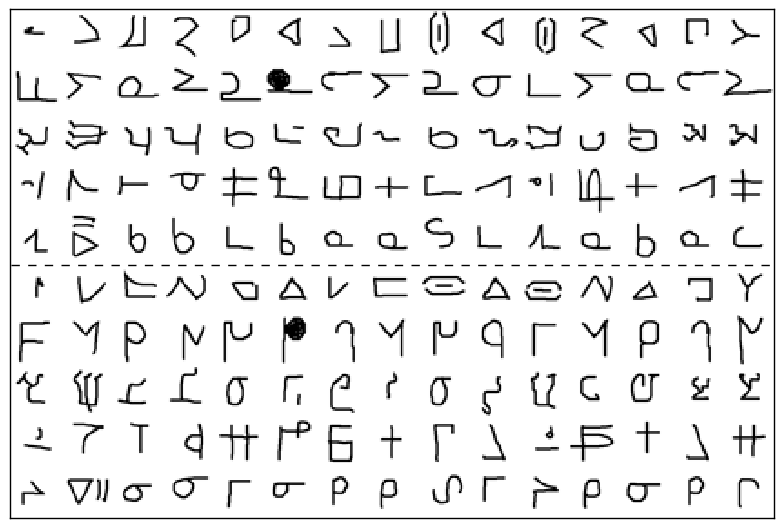}
\end{tabular}
\caption{\emph{Top:} Example characters from 5 alphabets taken from the Omniglot dataset.
\emph{Bottom:} Those same alphabets with each character rotated by $90^{\circ}$.
Distinguishing characters within each of these 10 alphabets constitute the 10 tasks in our multi-task learning experiment.}
\label{fig:omniglot_images}
\end{center}
\end{figure}

We use five alphabets from the Omniglot set. To see whether our multitask
learning system is able to learn high level similarities as well as
low-level similarities, we repeat these five alphabets with the images rotated
by 90 degrees (Figure \ref{fig:omniglot_images}) to make ten alphabets total.

Figure \ref{fig:omniglot_results} shows the learned penalties (normalized by row
and column to have ones on the diagonal, akin to a correlation matrix).
We see
that the lowest layer has been partially shared, across all alphabets equally,
with the upper layers much less shared. Interestingly, the top layer penalty
learns to share weights between the rotated alphabets.
\newcommand{\omniimagea}[2]{\parbox{4em}{\includegraphics[width=1.4cm]{minifigs/learned_corr_#1_#2.pdf}}}%
\newcommand{\omniimageb}[1]{\omniimagea{#1}{0} & \omniimagea{#1}{1} & \omniimagea{#1}{2}}%
\begin{figure}[h!]
\renewcommand{\tabcolsep}{1pt}
\begin{center}
\begin{tabular}{c@{\hskip 0.9em}ccc@{\hskip 0.9em}c@{\hskip 0.9em}c}%
& Input   & Middle  & Output & Train & Test\\
& weights & weights & weights & error & error \\
\parbox{3.7em}{Separate networks} & \omniimageb{no_sharing}      & 0.61 & 1.34\\ \hline
\parbox{3.7em}{Tied weights}      & \omniimageb{full_sharing}    & 0.90 & 1.25\\ \hline
\parbox{3.7em}{Learned sharing}   & \omniimageb{learned_sharing} & 0.60 & \bf 1.13
\end{tabular}
\caption{Results of the Omniglot multitask experiment.
Each matrix shows the degree of weight sharing between each pair of tasks for that layer.
\emph{Top}: A separate network is trained independently for each task.
\emph{Middle}: The lowest-level features were forced to be shared.
\emph{Bottom}: The degree of weight sharing between tasks was optimized by hyperparameter optimization.
}
\label{fig:omniglot_results}
\end{center}
\vskip -0.15in
\end{figure}

\subsection{Implementation Details}
Automatic differentiation (AD) software packages such as
Theano~\citep{Bastien-Theano-2012, bergstra2010scipy} are mainstays of deep
learning, significantly speeding up development time by providing gradients
automatically. Since we required access to the internal logic of RMD in order to implement Algorithm \ref{alg:reverse-sgd}, we implemented
our own automatic differentiation package for Python, available at \url{github.com/HIPS/autograd}.
This package differentiates standard
Numpy~\citep{oliphant2007python} code, and can differentiate code containing
while loops, branches, and even gradient evaluations.

Code for all experiments in this paper is available at \url{github.com/HIPS/hypergrad}.

\section{Limitations}

Back-propagation for training neural networks has several pitfalls that were later addressed by analysis and engineering.
Likewise, the use of hypergradients also has several apparent difficulties that need to be addressed before it becomes practical.
This section explores several issues with this technique that became apparent in our experiments.

\paragraph{When are gradients meaningful?}
\citet{bengio1994learning} noted that ``learning long-term dependencies with gradient descent is difficult.''
Our situation is even worse: We are using gradients to optimize functions which depend on their hyperparameters through hundreds of iterations of SGD.
To make things worse, each \primal{} iteration's gradient itself depends on forward- and then back-propagation through a neural network.
Thus the same issues that sometimes make \primal{} learning difficult are compounded.


For example, \citet*[Chapter 4]{pearlmutter1996investigation} showed that 
large learning rates induce chaotic behavior in the learning dynamics,
making the gradient uninformative about the medium-term shape of the training objective.
This phenomenon is related to the exploding-gradient problem~\cite{pascanu2012understanding}.

Figure \ref{fig:chaos} illustrates this phenomenon when training a neural network having 2 hidden layers for 50 \primal{} iterations.
\begin{figure}
\begin{center}
\includegraphics[width=0.9\columnwidth]{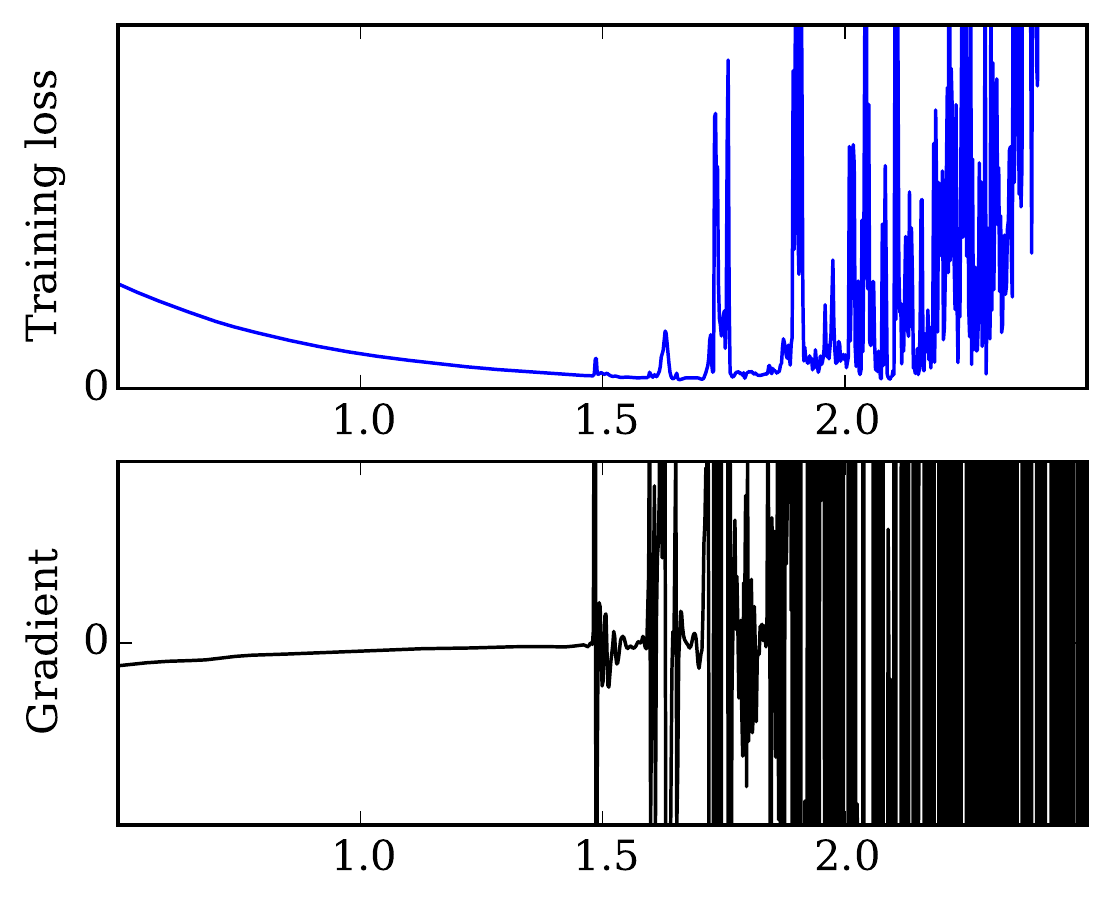}
\vskip -0.1in
\small Log learning rate
\caption{\emph{Top:} Loss after training as a function of learning rate.
\emph{Bottom:} Gradient of loss with respect to learning rate.
When the learning rate is high, the gradient becomes uninformative about the medium-term behavior of the function.
To maintain stability during meta-learning, we initialize using a small learning rate so as to approach the minimum from the left.}
\label{fig:chaos}
\end{center}
\vskip -0.2in
\end{figure} 
We partially addressed this problem in our experiments by initializing learning rates to be relatively small, and stopping meta-optimization when the magnitude of the meta-gradient began to grow.


\paragraph{Overfitting}
How many hyperparameters can we fruitfully optimize?
One limitation is overfitting the validation objective, in the same way that optimizing too many parameters can overfit the training objective.
However, the same rules of thumb still apply -- the size of the validation set, assuming examples are i.i.d., gives a rough guide to how many hyperparameters can be optimized.


\paragraph{Discrete parameters}
Of course, gradients are not necessarily useful for optimizing discrete hyperparameters such as the number of layers, or hyperparameters that affect discrete changes such as dropout regularization parameters.
Some of these difficulties could be addressed by parameterizing apparently discrete choices in a continuous manner.
For instance, the per-hidden-unit regularization of section \ref{sec:optimizing regularization} is an example of a continuous way to choose the number of hidden units.

\section{Related work}

The most closely-related work is \citet{domke2012generic}, who derived algorithms to compute reverse-mode derivatives of gradient descent with momentum and L-BFGS, using them to update the hyperparameters of CRF image models.
However, his approach relied on na\"ive caching of all parameter vectors $\params_1, \params_2, \dots, \params_T$, making it impractical for large models with many training iterations.

\citet{larsen1998adaptive}, \citet{eigenmann1999gradient}, \citet{chen1999optimal}, \citet{bengio2000gradient}, \citet{abdel2007adaptive}, and \citet{foo2008efficient} showed that gradients of regularization parameters are available in closed form when training has converged exactly to a local minimum.
In contrast, our procedure can compute exact gradients of any type of hyperparameter, whether or not learning has converged.

\paragraph{Support vector machines}
\citet{chapelle2002choosing} 
 introduced a differentiable bound on the SVM loss in order to be able to compute derivatives with respect to hundreds of hyperparameters, including weighting parameters for each input dimension in the kernel.
However, this bound was not tight, since optimizing the SVM objective requires a discrete selection of training points.

\paragraph{Bayesian methods}
For Bayesian models with a closed-form marginal likelihood, gradients with respect to all continuous hyperparameters are usually available.
For example, this ability has been used to construct complex kernels for Gaussian process models~\citep[Chapter 5]{rasmussen38gaussian}.
Variational inference also allows gradient-based tuning of hyperparameters in Bayesian neural-network models such as deep Gaussian processes~\citep{deepGPVar14}.
However, it does not provide gradients with respect to training parameters.


\paragraph{Gradients with respect to Markov chain parameters}
\citet{Bridging14} tune the step-size and mass-matrix parameters of Hamiltonian Monte Carlo by chaining gradients from a lower bound on the marginal likelihood through several iterations of leapfrog dynamics.
Because they used only a small number of steps, all intermediate values could be stored na\"ively.
Our reversible-dynamics memory-tape approach could be used to dramatically extend the number of HMC iterations used in this approach.

\section{Extensions and future work}

\paragraph{Bayesian optimization with gradients}
Hypergradients could be used with parallel, model-based optimization of hyperparameters.
For example, Gaussian-process-based optimization methods 
 could incorporate gradient information~\cite{solak2003derivative}.
Such methods could make use of parallel evaluations of hypergradients, which might be too slow to evaluate in a sequential manner.

\paragraph{Reversible \primal{} computation}
Recurrent neural network models can require so much memory to differentiate that checkpointing is required simply to compute their \primal{} gradients~\citep{martens2012training}.
Reversible computation might offer memory savings for some architectures.
For example, evaluations of Long Short-Term Memory~\citep{hochreiter1997long} or a Neural Turing Machines~\citep{graves2014neural} rely on long chains of mostly-small updates of parameters.
Exactly reversing these dynamics might allow more memory-efficient \primal{} gradient evaluations of their outputs on very long input sequences. 

\paragraph{Exactly reversing other learning methods}
The memory saving trick from Section \ref{sec:reversible computation} could presumably be applied to other momentum-based variants of SGD such as RMSprop~\cite{Tieleman2012} or Adam~\citep{Adam14}.

\section{Conclusion}

In this paper, we derived a computationally efficient procedure for computing gradients through stochastic gradient descent with momentum.
We showed how the approximate reversibility of learning dynamics can be used to drastically reduce the memory requirement for exactly back-propagating gradients through hundreds of training iterations.

We showed how these gradients allow the optimization of validation loss with respect to thousands of hyperparameters, something which was previously infeasible.
This new ability allows the automatic tuning of most details of training neural networks.
We demonstrated the tuning of detailed training schedules, regularization schedules, and neural network architectures.

\section*{Acknowledgments}
We would like to thank Christian Steinruecken, Oren Rippel, and Matthew James Johnson for helpful discussions.
We also thank Brenden Lake for graciously providing the Omniglot dataset.
Thanks to Jason Rolfe for helpful feedback.
We thank Analog Devices International and Samsung Advanced Institute of Technology for their support.

\section*{Appendix: Forward vs. reverse-mode differentiation}
\label{sec:appendix}
By the chain rule, the gradient of a set of nested functions is given by the product of the individual derivatives of each function:
\begin{align*}
\pderiv{f_4(f_3(f_2(f_1(x))))}{x} = \pderiv{f_4}{f_3} \cdot \pderiv{f_3}{f_2} \cdot \pderiv{f_2}{f_1} \cdot \pderiv{f_1}{x}
\end{align*}
If each function has multivariate inputs and outputs, the gradients are
Jacobian matrices.

Forward and reverse mode differentiation differ
only by the order in which they evaluate this product.
Forward-mode differentiation works by multiplying gradients in the same order as
the functions are evaluated:
\begin{align*}
\pderiv{f_4(f_3(f_2(f_1(x))))}{x} = \pderiv{f_4}{f_3} \cdot \left( \pderiv{f_3}{f_2} \cdot \left( \pderiv{f_2}{f_1} \cdot \pderiv{f_1}{x} \right) \right)
\end{align*}
Reverse-mode multiplies the gradients in the opposite order, starting from the
final result:
\begin{align*}
\pderiv{f_4(f_3(f_2(f_1(x))))}{x} = \left(  \left(  \pderiv{f_4}{f_3} \cdot \pderiv{f_3}{f_2} \right) \cdot \pderiv{f_2}{f_1} \right) \cdot \pderiv{f_1}{x} 
\end{align*}
In an optimization setting, the final result of the nested functions, $f_4$, is
a scalar, while the input $x$ and intermediate values, $f_1 - f_3$, can be
vectors. In this scenario the advantage of reverse-mode
differentiation is very clear. Let's imagine that the dimensionality of all the
intermediate vectors is $D$. In reverse mode, we start from the (scalar) output,
and multiply by the next $D \times D$ Jacobian at each step. The value we
accumulate is just a $D$-dimensional vector. In forward mode, however, we must
accumulate an entire $D \times D$ matrix at each step. But do we have still
have to compute and instantiate the $D \times D$ Jacobian matrices themselves
either way?  In general, yes. But in the (common) case that the vector-to-vector
functions are either elementwise operations or (reshaped) matrix multiplications, the
Jacobian matrices can actually be very sparse, and multiplication by the
Jacobian can be performed efficiently without instantiation~\cite{pearlmutter2008reverse}.

The main drawback of reverse-mode differentiation is that intermediate values
must be maintained in memory during the forward pass. In sections
\ref{sec:reversible learning} and \ref{sec:reversible computation}, we show how
to drastically reduce the memory requirements of reverse-mode differentiation
when differentiating through the entire learning procedure.

\bibliography{references.bib}
\bibliographystyle{icml2015stylefiles/icml2015}

\end{document}